\documentclass[sn-apa, iicol]{sn-jnl}% APA Reference Style 
\usepackage{graphicx}%
\usepackage{multirow}%
\usepackage{amsmath,amssymb,amsfonts}%
\usepackage{amsthm}%
\usepackage{mathrsfs}%
\usepackage[title]{appendix}%
\usepackage{xcolor}%
\usepackage{textcomp}%
\usepackage{manyfoot}%
\usepackage{booktabs}%
\usepackage{algorithm}%
\usepackage{algorithmicx}%
\usepackage{algpseudocode}%
\usepackage{listings}%
\usepackage{tabularray}
\usepackage{tabularx}
\usepackage{multirow}
\usepackage{subcaption}
\begin{document}

\title[A motion-based compression algorithm for resource-constrained video camera traps]{A motion-based compression algorithm for resource-constrained video camera traps}


\author*[1]{\fnm{Malika Nisal} \sur{Ratnayake}}\email{malika.ratnayake@monash.edu}

\author[1,2]{\fnm{Lex} \sur{Gallon}}\email{lex.gallon@monash.edu}

\author[3,4]{\fnm{Adel N} \sur{Toosi}}\email{adel.toosi@unimelb.edu.au}

\author[1]{\fnm{Alan} \sur{Dorin}}\email{alan.dorin@monash.edu}

\affil*[1]{\orgdiv{Agents and Decision-Making Group, Dept. of Data Science and AI, Faculty of Information Technology}, \orgname{Monash University}, \orgaddress{\city{Clayton}, \postcode{3800}, \state{Victoria}, \country{Australia}}}

\affil[2]{\orgdiv{Monash e-Research Centre}, \orgname{Monash University}, \orgaddress{\city{Clayton}, \postcode{3800}, \state{Victoria}, \country{Australia}}}

\affil[3]{\orgdiv{School of Computing and Information Systems, Faculty of Engineering and Information Technology}, \orgname{The University of Melbourne}, \orgaddress{\city{Parkville}, \postcode{3052}, \state{Victoria}, \country{Australia}}}

\affil[4]{\orgdiv{Dept. of Software Systems and Cybersecurity, Faculty of Information Technology}, \orgname{Monash University}, \orgaddress{\city{Clayton}, \postcode{3800}, \state{Victoria}, \country{Australia}}}

% \affil*[1]{\orgdiv{Department}, \orgname{Organization}, \orgaddress{\street{Street}, \city{City}, \postcode{100190}, \state{State}, \country{Country}}}

% \affil[2]{\orgdiv{Department}, \orgname{Organization}, \orgaddress{\street{Street}, \city{City}, \postcode{10587}, \state{State}, \country{Country}}}

% \affil[3]{\orgdiv{Department}, \orgname{Organization}, \orgaddress{\street{Street}, \city{City}, \postcode{610101}, \state{State}, \country{Country}}}


\abstract{Field-captured video facilitates detailed studies of spatio-temporal aspects of animal locomotion, decision-making and environmental interactions including predator-prey relationships and habitat utilisation. But even though data capture is cheap with mass-produced hardware, storage, processing and transmission overheads provide a hurdle to acquisition of high resolution video from field-situated edge computing devices. Efficient compression algorithms are therefore essential if monitoring is to be conducted on single-board computers in situations where such hurdles must be overcome. Animal motion tracking in the field has unique characteristics that necessitate the use of novel video compression techniques, which may be underexplored or unsuitable in other contexts. In this article, we therefore introduce a new motion analysis-based video compression algorithm specifically designed for camera traps. We implemented and tested this algorithm using a case study of insect-pollinator motion tracking on three popular edge computing platforms. The algorithm identifies and stores only image regions depicting motion relevant to pollination monitoring, reducing overall data size by an average of 87\% across diverse test datasets. Our experiments demonstrate the algorithm’s capability to preserve critical information for insect behaviour analysis through both manual observation and automatic analysis of the compressed footage. The method presented in this paper enhances the applicability of low-powered computer vision edge devices to remote, \textit{in situ} animal motion monitoring, and improves the efficiency of playback during behavioural analyses. Our new software, \textit{EcoMotionZip}, is available Open Access.}




\maketitle

\clearpage
\section{Introduction}\label{sec1}

Environmental change and other anthropogenic factors are increasingly affecting wildlife, making fauna monitoring crucial to help manage or mitigate impacts. Camera traps placed in the wild have emerged as indispensable tools for ecologists to study animal behaviour \citep{caravaggi2017review}, habitat utilisation \citep{head2012remote, lovell2022effect, dharmarathne2022camera}, and species abundance and distributions \citep{reece2021camera, feng2021assessing}. Currently, the applicability of camera traps, especially those that collect high resolution video, is limited by transmission bandwidth constraints, particularly in remote areas where data is transferred over wireless cellular or satellite networks \citep{wong2024camera}. Therefore, enhancements in data compression, processing, and transmission, are required to capitalise on the availability of cheap, robust data capture devices. In this article, we explore this general problem by presenting a set of properties specific to the application of data-rich video camera traps to wildlife monitoring. The specificity of these needs suggests avenues to explore for novel solutions to improve monitoring device autonomy that are arguably more important for this domain than in the domains dominating algorithm development, such as closed-circuit television (CCTV) for security and surveillance. From this high-level perspective we address a particularly challenging case study, the design of a novel system for video monitoring pollinator insects.

The unpredictability of both the outdoor environment and the animals being observed makes wildlife monitoring with camera traps particularly challenging. Wildlife monitors must withstand extreme weather conditions, temperature fluctuations, and potential vandalism from both humans and animals \citep{meek2022mitigating}. Also, wildlife monitoring often involves long periods of inactivity, punctuated by bursts of animal activity that can be difficult to predict or capture. This variability is compounded by the difficulty in distinguishing between individual animals within a species, especially those with similar size and morphological features. This challenge is particularly acute for monitoring fish in schools or insects in swarms, where individual identification may increase understanding of how individual behaviours contribute to population or group level dynamics and behaviour. The physical environment also presents significant obstacles for wildlife monitoring. The need to survey a wide field of view in the wild to capture animal movement conflicts with the requirement for high-detail, focused data to identify and track individual animals, especially small, fast-moving insects \citep{ratnayake2023spatial}. Also, the visual complexity and concealment provided by some natural backgrounds can make it difficult to reliably detect and track moving animals \citep{moll2020effect}.  Finally, environmental variations, such as the arrival or loss of flowers, water, shade etc., influence animal movements and behaviour. This requires camera traps to maintain an accurate record of the environment \textit{and} animals to inform analysis. This necessitates the considered use of data capture techniques and equipment that can adapt to and capture changing conditions while minimising disturbance to the animals being observed.

Wildlife camera trap data can be collected as still images or video. Still image traps capture single or short-burst sequences of a location at predetermined intervals, or when triggered by animals \citep{collett2017time}. Video camera traps may record continuous, sometimes lengthy, image sequences at high sampling rates (e.g. $30+$ frames/sec), or at preset intervals, or when triggered. Video data allows analysis of dynamic interactions, such as bird courtship displays \citep{janisch2021video}, predator ambush strategies \citep{rampim2020antagonistic}, and animal pollination \citep{krauss2018effectiveness, ratnayake2023spatial, melidonis2015diurnal} that aren't fully captured in still images.  However, a video camera trap's rich data comes at a cost: video storage, processing, and transmission from remote devices with limited access to power, storage capacity, and transmission bandwidth, are challenging. Some systems, therefore, run only for a few hours or days at a stretch, avoid onboard-video processing, and require manual data transfer \citep{droissart2021pict}. This can reduce the value of a video camera trap. Ideally, it would be capable of remote autonomous operation with infrequent service visits, and all data would be streamed conveniently to wildlife management and research offices. It is this need for increased autonomy of video camera traps that drives our research. We address the challenge of limited storage and transmission bandwidth by compressing video camera data files with a new algorithm tailored specifically for wildlife monitoring. Insect pollinator monitoring is our case study to demonstrate algorithm performance.

Monitoring insect pollinators is valuable to support them in sustaining natural ecosystems \citep{fao2019} and global food production \citep{gazzea2023global}. However, observing insects outdoors is difficult due to the complexities of their environment and the insects' own varied appearances and behaviours \citep{nykanen2023motion}.  The natural environments where insects forage are subject to frequent change due to wind and sunlight, making it difficult to accurately track their movement. Also, insect species vary widely in size, morphology, microhabitat preferences, and motion characteristics, all of which add to the difficulty of recording them. Video capture is valuable in such studies to provide high spatio-temporal resolution data for insect/environment interaction monitoring. 

As noted above, current video-based insect camera traps employ continuous recording \citep{droissart2021pict, ratnayake2023spatial, ratnayake2021tracking}, time-lapse \citep{naqvi2022camera}, or motion-triggers  \citep{van2022continuous, naqvi2022camera}. Continuous video recordings require high storage capacity and transmission bandwidth. Time-lapse videos require less storage and bandwidth, but miss activity during non-recording periods. Motion-triggering aims to reduce storage and transmission requirements by recording video only when insects are in view. However, widely-used hardware triggers like (Passive Infra-Red) PIR sensors, are ineffective for detecting small insects \citep{ortmann2021reliable, naqvi2022camera}. Software-based motion triggers have also been implemented utilising foreground-background segmentation \citep{van2022continuous} but are susceptible to false positive detections caused by wind moving foliage or illumination changes. Deep learning \citep{sittinger2024insect} triggers can minimise false positives by recording videos only when an insect is in the camera frame. However, these models usually require substantial computational resources, specialised hardware, and detection models trained on a wide variety of species. This can limit their value for autonomous applications and makes them prone to mis-detections.

In this paper, we introduce a novel approach that substantially compresses videos without compromising insect data. Our method processes camera data frame-wise and pixel-wise to identify image regions with motion, storing only the information relevant for animal monitoring and reconstruction of motion paths and animal-habitat (in our case, flower) interaction. The algorithm is designed to accommodate limited power, storage, and bandwidth resources of camera traps installed in remote locations while improving processing throughput. Moreover, it can be seamlessly integrated with existing video data collection methods, including continuous recording, time-lapse, and motion-triggered recording, to improve their overall performance. In addition, we extend the existing open-source outdoor insect tracking software Polytrack \citep{ratnayake2023spatial} with the new algorithm, enabling it to automatically track insects in compressed videos to extract motion trajectories. We report on a case study applying our algorithm to six datasets representing diverse application environments. We demonstrate the value of our method by extracting insect behavioural data from compressed videos. Alongside this article, we publish the compression algorithm (\textit{EcoMotionZip}), implementation documentation, and a modification of Polytrack.

\section{Materials and Methods}\label{sec2}

\subsection{Video compression algorithm}

\begin{figure*}[h!]
    \centering
    \includegraphics[width = \linewidth]{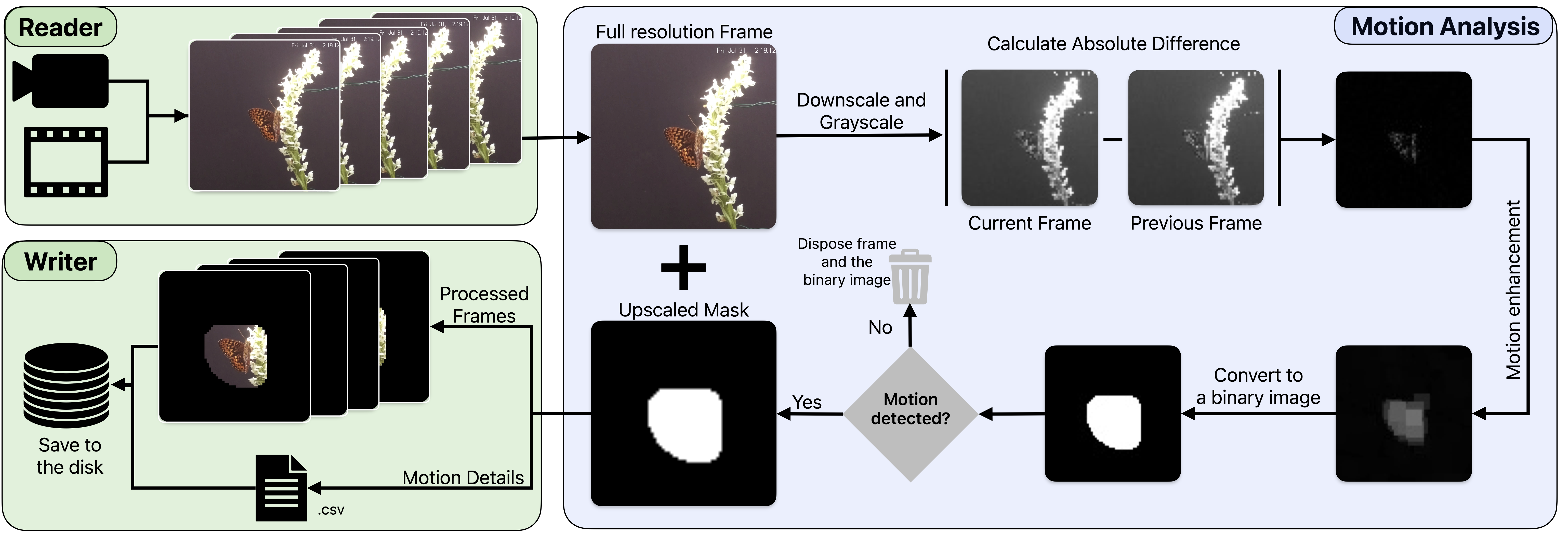}
    \caption{\textbf{Overview of the algorithm architecture.} The proposed algorithm has three main components: (1) Reader, (2) Motion Analysis, and (3) Writer, implemented with three separate threads. Footage from \citet{van2022continuous} illustrates the algorithm.}
    \label{fig:method_overview}
\end{figure*}

In this section we describe the algorithm's multi-threaded architecture (Fig.\ref{fig:method_overview}) designed to improve data throughput. 

\subsubsection{Reader component}

The Reader captures video frames from a camera stream or pre-recorded video file and adds them to a queue for processing. Once all frames from the input stream have been passed to the Motion Analysis component for processing, the Reader terminates.

% The reader captures video frames from either a camera stream or a pre-recorded video file and adds them to a queue for processing. This multi-threaded approach enhances the algorithm's efficiency by preventing frame drops caused by computationally intensive processes. The reader thread operates at a higher frame rate than the camera's capture rate (30 fps) to ensure a sufficient buffer for the motion analysis thread. Users can choose between real-time camera capture or a pre-recorded video as the input source. Once all frames from the input stream have been passed to the motion analysis thread for processing, the reader thread terminates.

\subsubsection{Motion Analysis component}

The Motion Analysis component processes frames captured by the \textit{Reader} thread to identify regions with motion. It is designed to extract only information from video frames critical for animal behaviour analysis. It discards the remaining information to save storage space. Maintained data includes information for algorithms (and human viewers) to gauge (1) animal type / species, (2) movement paths / gaits, (3) observation time, and (4) a snapshot of the environment within the camera view.

Motion analysis begins by down-scaling each captured frame and converting it to greyscale to reduce computational load and improve processing efficiency \citep{ratnayake2021towards,bjerge2023object}. Subsequently, inter-frame changes are detected by calculating the absolute intensity difference between pixels in adjacent frames. Pixel regions displaying an absolute intensity difference greater than a user-set threshold are preserved and expanded to include a surrounding buffer region. Users can adjust the sensitivity of the motion capture by modifying the threshold value to suit the monitored environment and target species. The buffer region allows the algorithm to maintain detailed information about the animal and its immediate surroundings, facilitating subsequent behavioural analysis. The size of the buffer region can be customised by the user. Subsequently, the frame containing motion regions is converted into a binary image, where pixels of non-zero intensity are retained while the rest are zeroed. Frames with no regions of motion are discarded completely. Frames with regions of detected motion undergo upscaling of the binary image to the original frame size and a bitwise product is generated between the upscaled binary image and the original frame. The resultant frame is then passed to the Writer component for storage. The Motion Analysis component terminates once all frames from the Reader have been processed. In addition, the algorithm captures full frames at the start and at user-specified intervals during motion sequences to provide an overview of the scene and document any gradual changes in the environment that occur over the recording period. Alongside processed motion frames sent to the Writer, the Motion Analysis component transmits the frame numbers of the input and output videos, and whether or not a full image frame has been saved. This information is stored in a CSV file for later reconstruction of animal motion and behaviour.% In addition, the algorithm records full frames at start and during motion sequences at user-specified intervals to capture a scene overview and record any changes in the environment that may have occurred gradually during the recording period.
% (regardless of whether or not motion was detected in that frame)

\subsubsection{Writer component}

The \textit{Writer} component receives processed frames from the Motion Analysis component and re-assembles them into a video file. The Writer extracts the frame rate and resolution of the output video from the input video file. Additionally, the Writer stores a CSV file alongside the video file containing the supplementary data sent by the Motion Analysis component.

\subsubsection{Software implementation}

We have implemented the proposed algorithm as \textit{EcoMotionZip} available on GitHub via the link provided under the Code Availability section. EcoMotionZip was developed using Python 3.11.2, Computer Vision Library (OpenCV) 4.6.0, Numpy 1.24.2, and FFMPEG 5.1.4. In addition to processing and compressing pre-recorded videos, the software supports real-time video recording and processing with the PiCamera2 library. Detail on software dependencies, installation and user guides appear on GitHub.

\subsection{Behavioural analysis}

\begin{figure*}[h!]
    \centering
    \includegraphics[width = \linewidth]{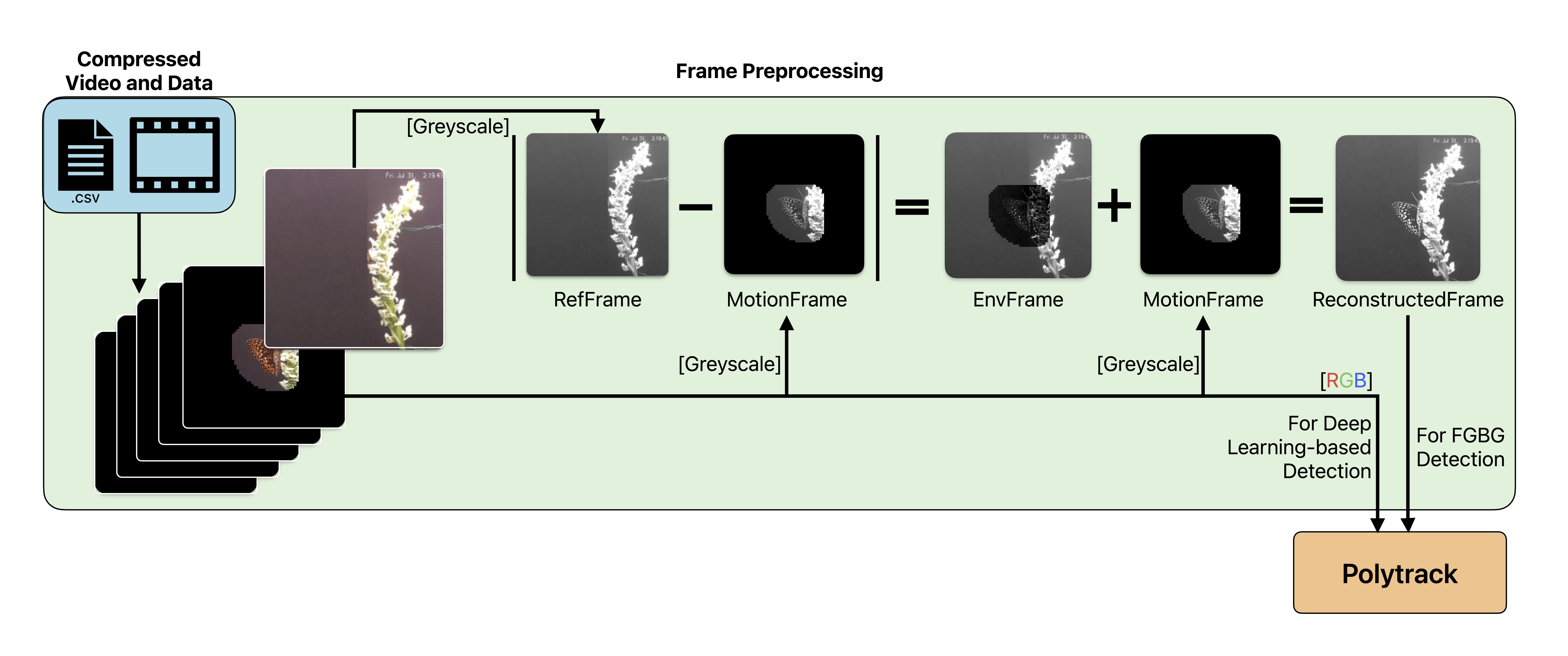}
    \caption{\textbf{Frame pre-processing for compressed videos}. This step processes video frames captured using a compression algorithm to make them compatible with Polytrack analysis \citep{ratnayake2023spatial}. ``For Deep Learning-based Detection" indicates the RGB image input to Polytrack to be used by the deep learning-based detection model and ``For FGBG Detection" shows the image input to Polytrack to be used by the foreground-background segmentation model.}
    \label{fig:polytrack_overview}
\end{figure*}

The ability to automatically process camera trap data for behavioural information is crucial for analysing data collected over long periods. While current software options are limited, Polytrack software \citep{ratnayake2023spatial} has shown promise in tracking multiple insect species simultaneously and monitoring their flower visiting behaviour outdoors. However, Polytrack was originally designed for uncompressed video. We modified Polytrack by incorporating a frame pre-processing step to handle compressed videos. This modification allows for the extraction of insect trajectories using information from compressed videos and additional details provided in separate CSV files. Fig.\ref{fig:polytrack_overview} displays an overview of the frame pre-processing step incorporated into Polytrack.

% \subsubsection{Frame pre-processing}
Polytrack employs a hybrid detection and tracking model \citep{ratnayake2021tracking} that combines a deep learning-based object detector with a foreground/background (FGBG) segmentation model to track insects and flowers. However, motion regions captured in compressed videos from EcoMotionZip often include a buffer region showing the animal's immediate surroundings. This can reduce the accuracy of insect bounding box calculations made by the FGBG segmentation detector in Polytrack. The objective of the frame pre-processing step is to eliminate the buffer region, allowing Polytrack's FGBG segmentation model to calculate a bounding box that tightly fits around the insect's body, thereby improving detection precision. Pre-processing starts by retrieving information from the CSV file about the positions of full frames in the video sequence, where snapshots of the entire scene are captured. Subsequently, frames are sequentially extracted from the video, and the full frames are stored as reference frames after their conversion to grayscale. Frames containing motion data are then processed with the stored reference frames, as outlined in Equations \ref{eqn:env_frame} and \ref{eqn:reconstructed_frame}, to obtain the respective reconstructed frames with both motion and environmental pixels. Finally, these reconstructed frames are transferred to Polytrack along with the original compressed video frames for insect tracking and behaviour analysis.%The objective of the frame pre-processing step is to eliminate the buffer region to improve object detection with Polytrack's FGBG segmentation model in Polytrack. This pre-processing process begins with retrieving information about the positions of full frames from the CSV files. 
\begin{equation}
    EnvF(x,y) = \left | RefF(x,y) - MotF(x,y) \right |
    \label{eqn:env_frame}
\end{equation}
\begin{equation}
    RecF(x,y) = EnvF(x,y) + MotF(x,y)
    \label{eqn:reconstructed_frame}
\end{equation}
\noindent where, $RefF$ if the reference frame (full frame), $MotF$ is the compressed frame with motion data, $EnvF$ is the resulting frame representing the absolute difference between the corresponding pixel values of the $RefF$ and $MotF$, and $RecF$ is the resultant frame with motion and environment information. $(x,y)$ denotes the coordinates of a pixel in the frame.

\subsubsection{Software implementation}

We modified Polytrack to incorporate a frame pre-processing step for compressed videos. This is available open-source on GitHub via the link provided under the Code Availability section. The deep learning object detection model used within Polytrack is Ultralytics YOLOv8 \citep{Jocher_Ultralytics_YOLO_2023}, trained on a subset of the annotated dataset published in \citep{Ratnayake2022_dataset}.

\subsection{Test datasets}

\begin{figure*}
    \centering
    \begin{subfigure}{0.3\textwidth}
        \includegraphics[width=\linewidth]{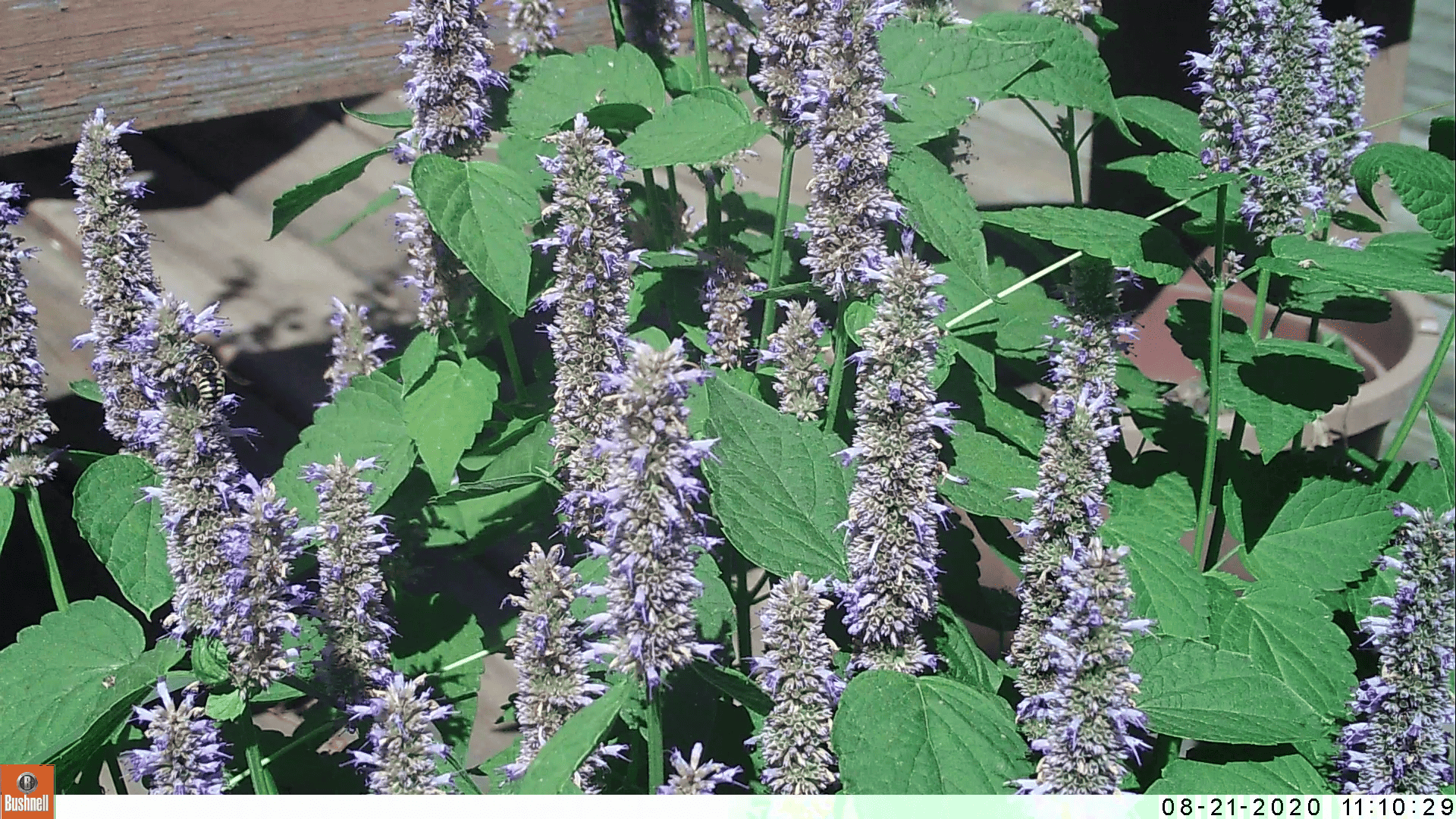}
        % \label{fig:subfig1}
        \caption{}
    \end{subfigure}
    \hfill
    \begin{subfigure}{0.3\textwidth}
        \includegraphics[width=\linewidth]{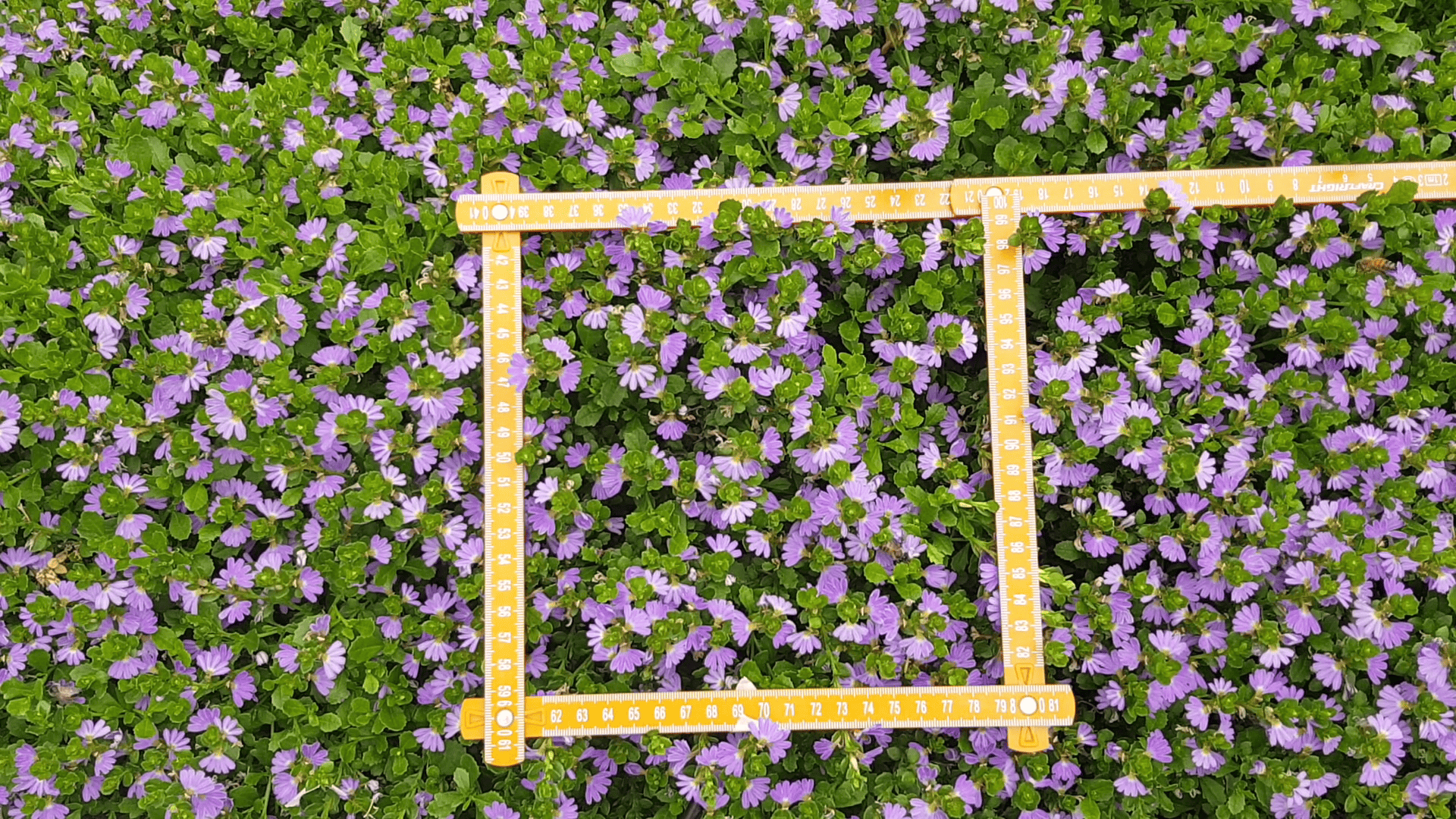}
        % \label{fig:subfig2}
        \caption{}
    \end{subfigure}
    \hfill
    \begin{subfigure}{0.3\textwidth}
        \includegraphics[width=\linewidth]{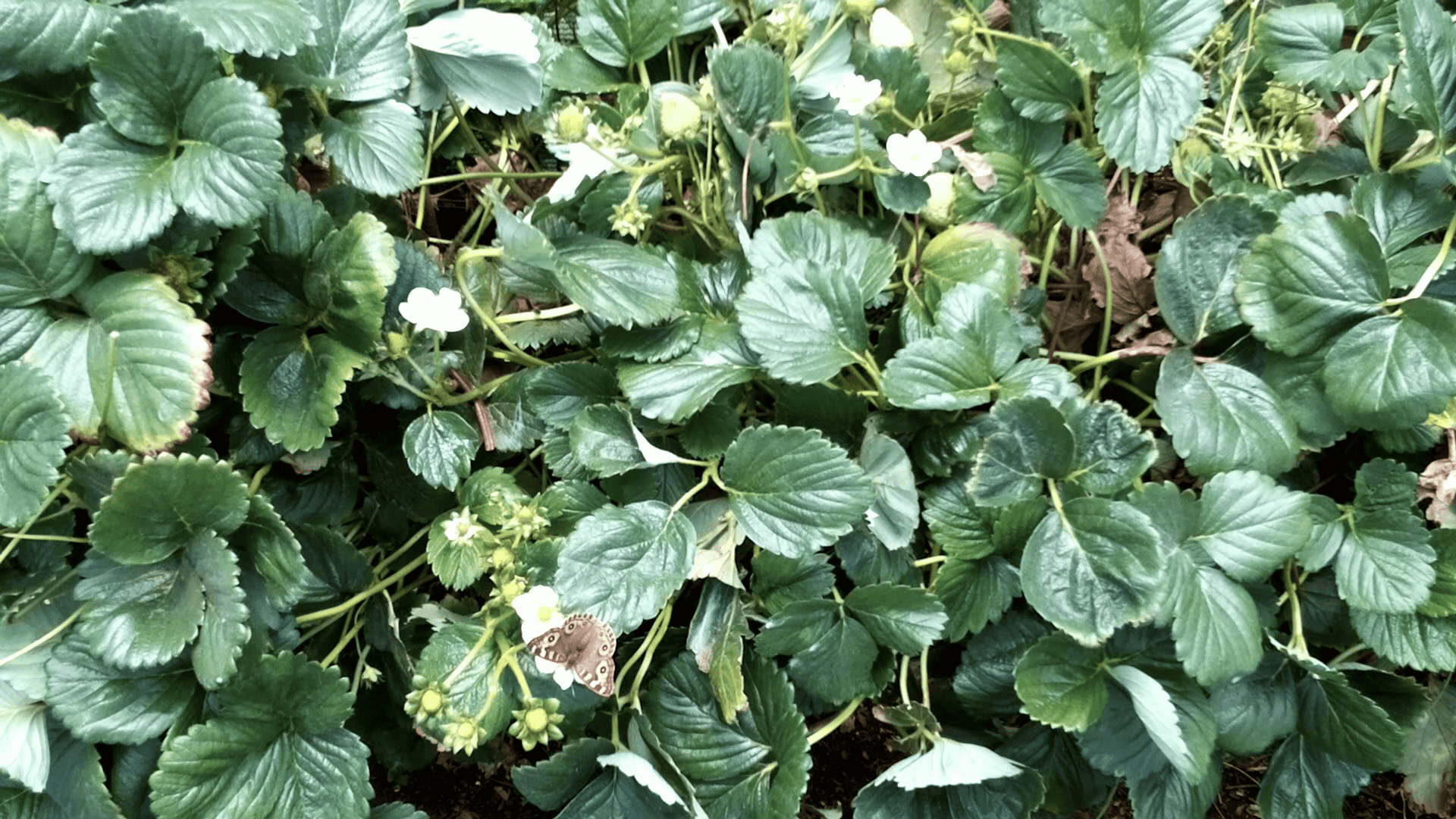}
        % \label{fig:subfig3}
        \caption{}
    \end{subfigure}
    \vspace{0.75cm}
    \begin{subfigure}{0.3\textwidth}
        \includegraphics[width=\linewidth]{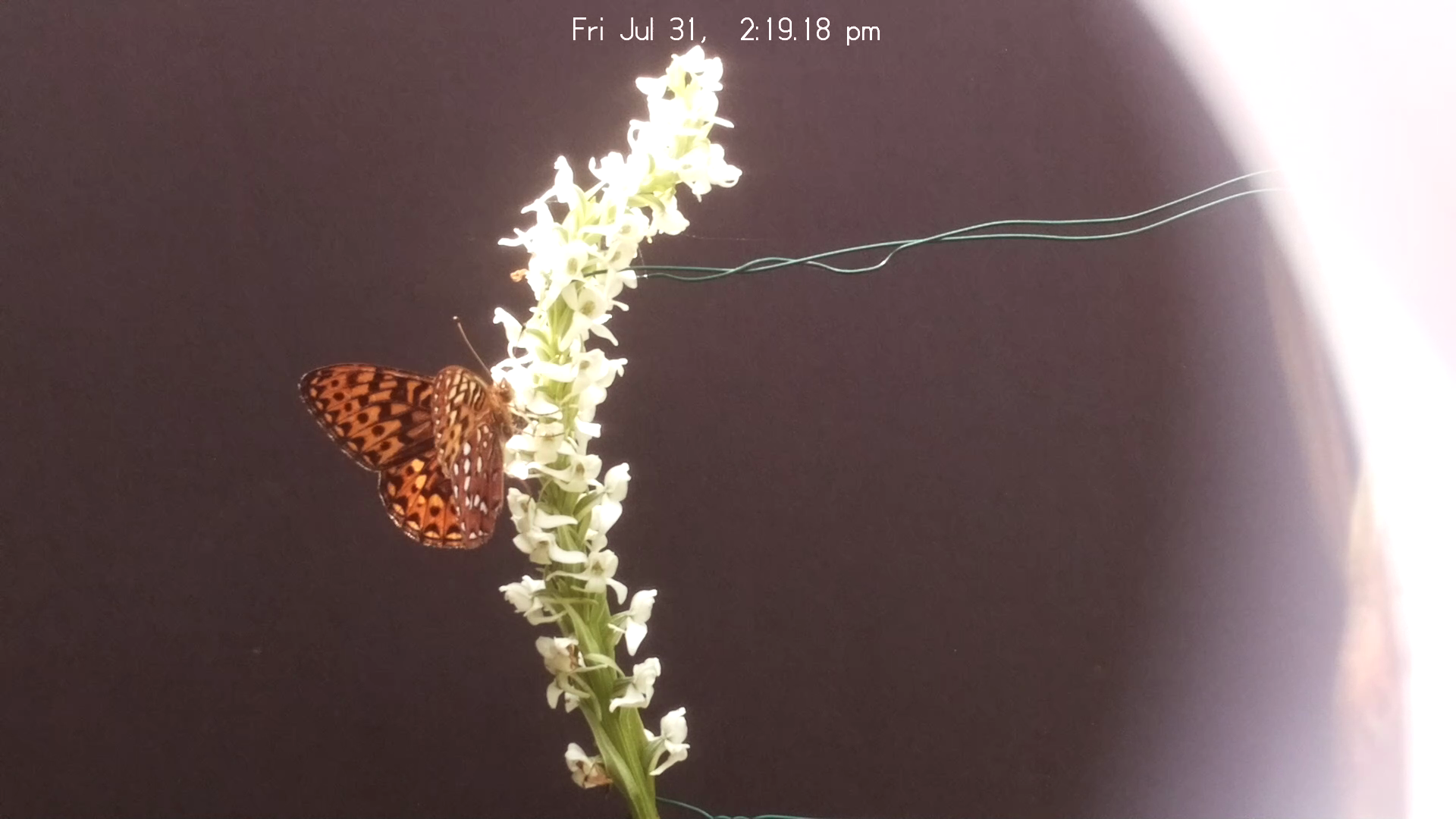}
        % \label{fig:subfig4}
        \caption{}
    \end{subfigure}
    \hfill
    \begin{subfigure}{0.3\textwidth}
        \includegraphics[width=\linewidth]{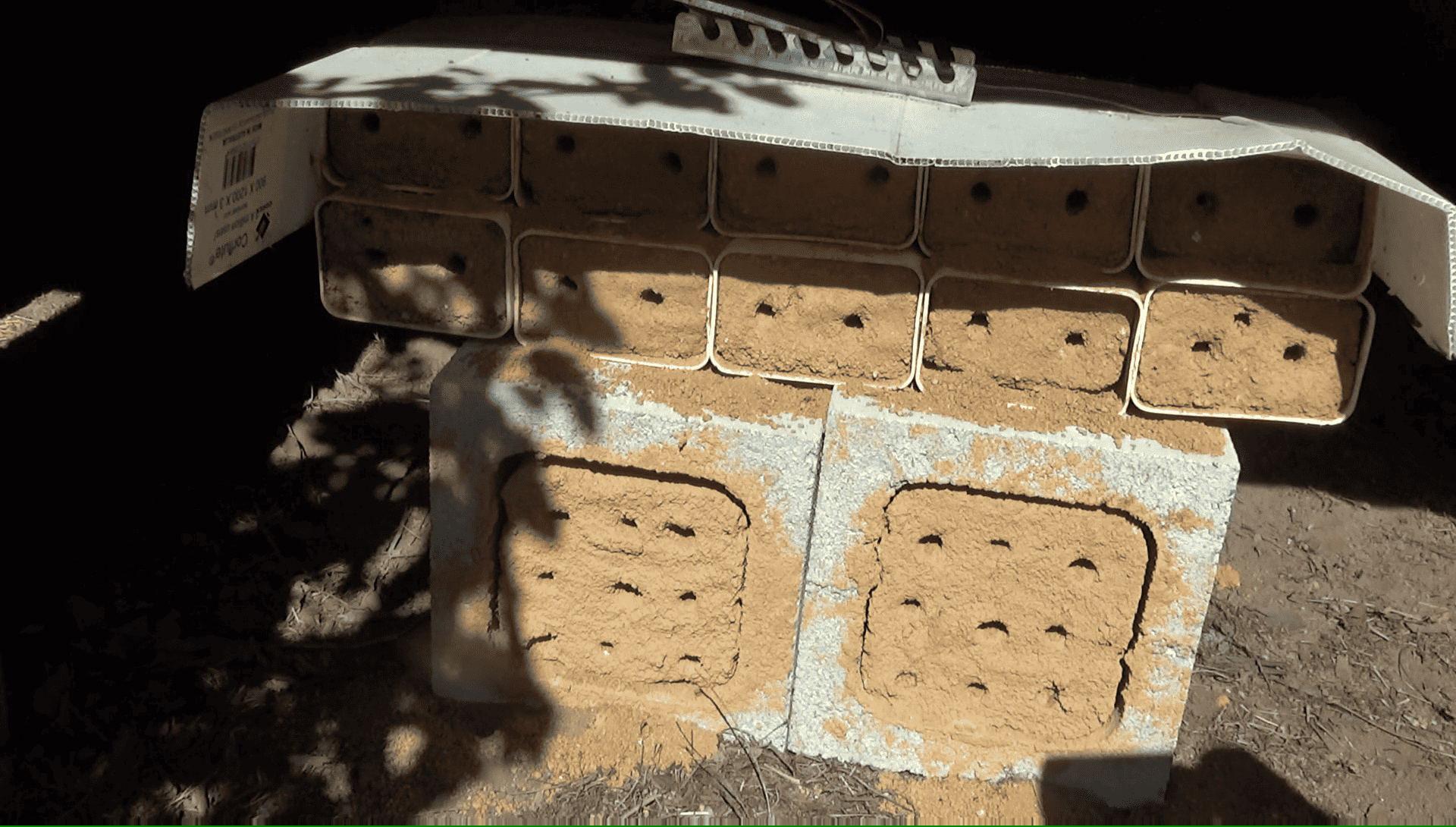}
        % \label{fig:subfig5}
        \caption{}
    \end{subfigure}
    \hfill
    \begin{subfigure}{0.3\textwidth}
        \includegraphics[width=\linewidth]{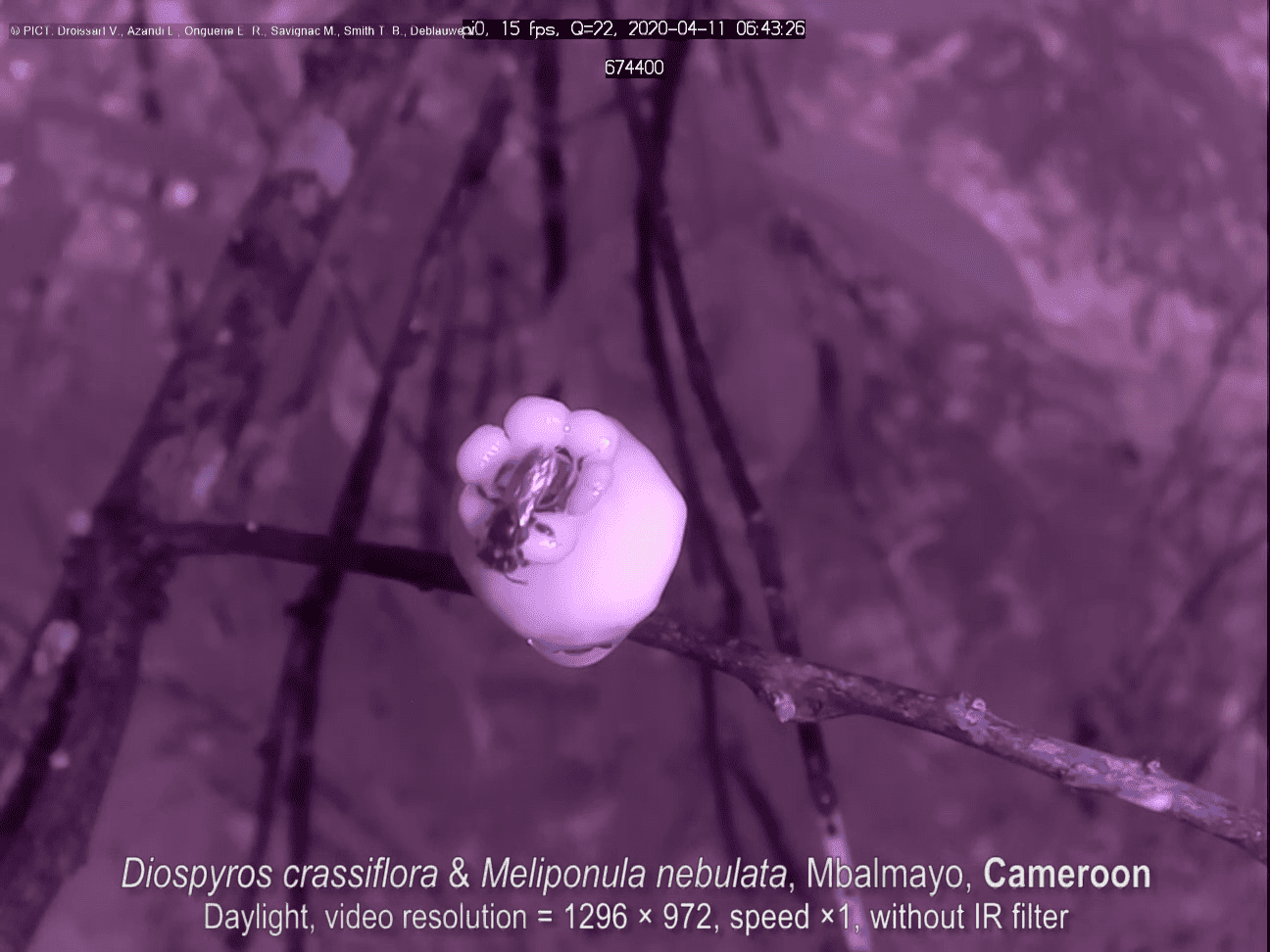}
        % \label{fig:subfig6}
        \caption{}
    \end{subfigure}
    \caption{\textbf{Application environments in the test dataset.} (a) \cite{naqvi2022camera}, (b) \cite{Ratnayake2020_dataset}, (c) \cite{Ratnayake2022_dataset}, (d) \cite{van2022continuous}, (e) Nest Monitoring, and (f) \cite{droissart2021pict}}
    \label{fig:application_envionments}
\end{figure*}

\begin{figure*}
\centering
    \includegraphics[width = \linewidth]{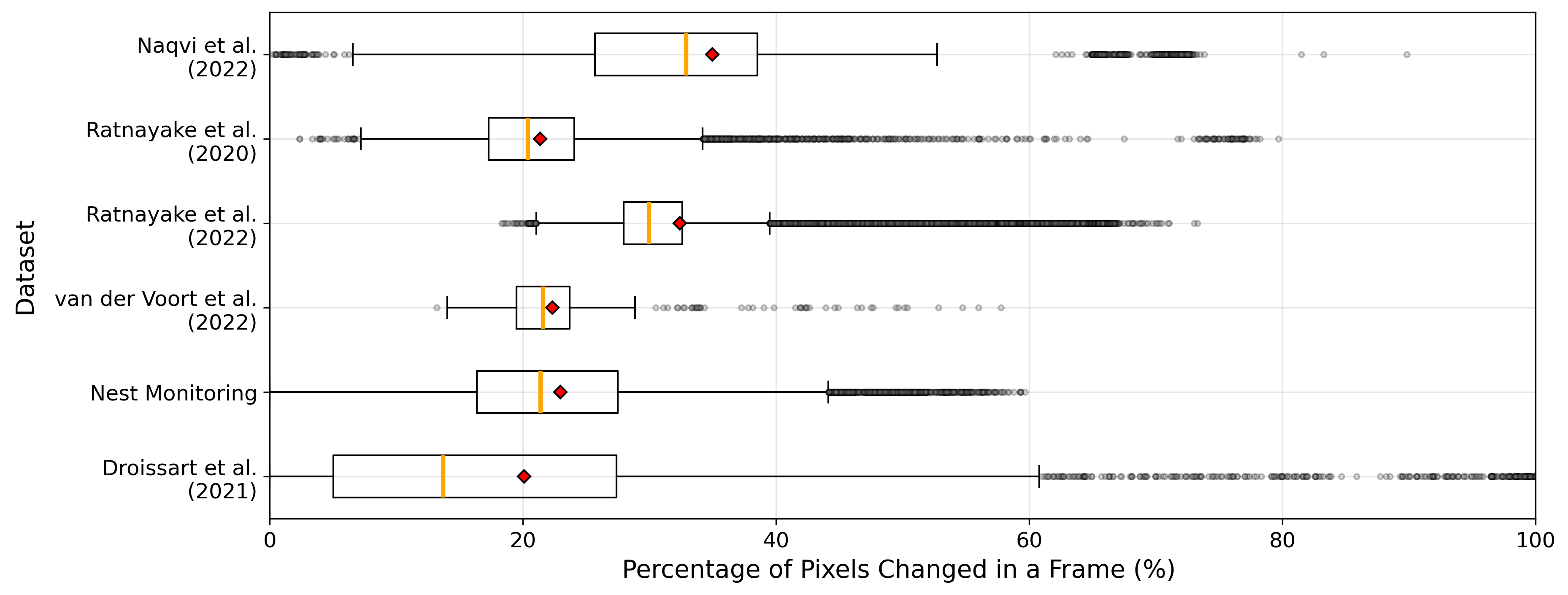}
    \caption{\textbf{Percentage of pixels changed per frame in test datasets}. The percentage of pixels changed per frame represents the proportion of pixels with altered intensity values between two consecutive frames in the video dataset. The red diamonds indicate the mean values, and the orange line represents the median for each dataset.}
    \label{fig:pixel_change}
\end{figure*}

To assess our video compression algorithm we used six real-world datasets of multiple videos encompassing a range of insect monitoring scenarios, application contexts, scene complexities, and recording modes (Table \ref{tab:video_datasets}). Fig.\ref{fig:application_envionments} shows snapshots of application environments and Fig.\ref{fig:pixel_change} shows the percentage pixels changed per frame recorded in datasets.

\begin{table*}
\centering
\caption{\textbf{Test video dataset information.}``Application Environment'' describes the monitoring environment and plant species recorded in the videos. ``Recording Method'' presents the video capture method. ``Camera'' describes the camera model used, ``No. Videos'' shows the number of videos in each dataset, ``Video Codec'' is the codec used, ``Video Resol.'' is the video resolution, ``FPS'' is the recording frame rate.}
\small
\label{tab:video_datasets}
\begin{tblr}{
  width = \linewidth,
  colspec = {Q[131]Q[192]Q[144]Q[182]Q[67]Q[71]Q[102]Q[42]},
  cells = {c},
  hline{1-2,8} = {-}{},
}
\textbf{Dataset} & {\textbf{Application}\\\textbf{Environment}} & {\textbf{Recording}\\\textbf{Method}} & \textbf{Camera} & {\textbf{No.~}\\\textbf{Videos}} & {\textbf{Video}\\\textbf{Codec}} & {\textbf{Video}\\\textbf{Resol.}} & \textbf{FPS}\\
\cite{naqvi2022camera} & {Urban garden \\(Mixed native and \\exotic species)} & {Motion triggered / \\ Time Lapse} & {Bushnell NatureView \\HD 119740} & 3 & H264 & 1920, 1080 & 30\\
\cite{Ratnayake2020_dataset} & {Urban garden\\~(Scaevola sp. native)} & Continuous & Samsung Galaxy S8 & 7 & H264 & 1920, 1080 & 60\\
\cite{Ratnayake2022_dataset} & {Commercial farm \\(Fragaria × ananassa, \\Strawberry crop)} & Continuous & Raspberry Pi V2 & 10 & MPEG4 & 1920, 1080 & 30\\
\cite{van2022continuous}  & {Controlled \\environment \\(Orchidaceae sp.)} & Motion~\textcolor[rgb]{0.125,0.129,0.141}{triggered} & Raspberry Pi V2 & 1 & H264 & 1920, 1080 & 24\\
Nest Monitoring & {Bee nest \\(Amegilla sp.)} & Continuous & Sony Handycam HDR-CX405 & 3 & H264 & 1920, 1080 & 25\\
\cite{droissart2021pict} & {Semi-controlled \\environment \\(Multiple sp.)} & Continuous & Raspberry Pi V2 & 3 & H264 & {1296, \\ 972} & -
\end{tblr}
\begin{tablenotes}
    \small
    \item {$^{\ast}$ \cite{droissart2021pict} dataset contains videos with different frame rates.}
\end{tablenotes}
\end{table*}

\section{Results}\label{sec3}

We conducted experiments evaluating performance of our algorithm and its ability to retain information critical for automated tracking and manual insect behavioural analysis. In this section we present the results of these experiments. A detailed description of experimental results is available in Supplementary Information.

\subsection{Video compression}

We evaluated video compression performance by comparing the reduction in file size and frame count. We maintained the video resolution, frame rate and video codec of the original video to clearly evaluate the compression capabilities of our algorithm. The test data was processed using Raspberry Pi 5 (8 GB) for the analysis. Results are shown in Table \ref{tab:video_compression}.

\begin{table*}[h]
\centering
\caption{\textbf{Results of the video compression.} ``No. Frames'' and ``File Size (MB)'' data for raw and processed videos show dataset total frame counts and file sizes. Reported file sizes for processed videos are the totals of compressed video and CSV files storing video supporting data. ``Frame Reduc. (\%)'' and ``File Size Reduc. (\%)'' show reduction in total frame count and test video file size.}
\label{tab:video_compression}
\small
\begin{tblr}{
  width = \linewidth,
  colspec = {Q[146]Q[115]Q[136]Q[115]Q[136]Q[140]Q[140]},
  cells = {c},
  cell{1}{1} = {r=2}{},
  cell{1}{2} = {c=2}{0.251\linewidth},
  cell{1}{4} = {c=2}{0.251\linewidth},
  cell{1}{6} = {r=2}{},
  cell{1}{7} = {r=2}{},
  hline{1,3,9} = {-}{},
  hline{2} = {2-5}{},
}
\textbf{Dataset } & \textbf{Raw Videos } &  & \textbf{Processed Videos } &  & {\textbf{Frame~}\\\textbf{Reduc. }~\textbf{(\%)}} & {\textbf{File Size }\\\textbf{Reduc. (\%)}}\\
 & \textbf{No. Frames} & \textbf{File Size (MB)} & \textbf{\textbf{No. Frames}} & \textbf{File Size~(MB)} &  & \\
 \cite{naqvi2022camera} & 5445 & 327.48 & 4166 & 70.61 & 23.49 & 78.44\\
\cite{Ratnayake2020_dataset} & 22269 & 442.45 & 9989 & 39.48 & 55.14 & 91.08\\
\cite{Ratnayake2022_dataset} & 179912 & 10895.05 & 12423 & 266.02 & 93.03 & 97.56\\
\cite{van2022continuous}  & 790 & 23.59 & 775 & 2.47 & 1.90 & 89.53\\
Nest Monitoring & 56664 & 1303.21 & 14331 & 52.92 & 74.71 & 95.94\\
\cite{droissart2021pict} & 5471 & 73.09 & 3862 & 20.80 & 29.41 & 71.54
\end{tblr}
\end{table*}

The proposed algorithm achieved an average of 87\% video compression across all test datasets. The minimum recorded compression was 71\% for the \cite{droissart2021pict} dataset. The compression ratio was particularly high for datasets where the environment remained relatively stationary during data collection, such as those in \cite{Ratnayake2020_dataset, Ratnayake2022_dataset, van2022continuous}. Notably, for all test videos, the file size reduction exceeded the frame count reduction. This difference was much more prominent for datasets recorded with motion-based triggers \citep{naqvi2022camera, van2022continuous}. This suggests that pixel-wise analysis of video frames, compared to simple motion detection, can achieve a higher compression rate (see Discussion).

\subsection{Information retention}

We evaluated our algorithm's ability to preserve relevant animal behaviour information for our case study by comparing the number of insect appearances detected in raw and compressed videos using both manual and automated techniques. We used the dataset \citet{Ratnayake2022_dataset} for this experiment as it has the highest video compression and hence, plausibly, the highest likelihood of information loss. This dataset contains video of four insect types: honeybees, Syrphid flies, Lepidopterans, and Vespids. We followed the procedure in \cite{ratnayake2023spatial} to record insect events in this dataset.

\subsubsection{Manual video observations}

 We conducted a frame-by-frame analysis of the compressed video footage and manually counted the number of insects, categorised them by type, and recorded the number of flowers visited by each insect. When an insect first appeared in the video, we played the footage frame-by-frame to analyse its movement. If an insect exited the frame but subsequently reappeared, or if it was occluded by foliage and re-emerged, we counted it as a new individual. Finally, when an insect landed on a fully visible and completely open flower, or if the compressed video had a gap in recorded frames when an insect was over a flower, we considered and recorded this as a flower visit. We only considered insects that appeared for more than 5 frames for the analysis, following the procedure outlined in \cite{ratnayake2023spatial}. We compared the results of these observations against the visual observation results published in the study. Results are shown in Table~\ref{tab:behaviour_analysis}.

\begin{table*}
\centering
\caption{\textbf{Comparison of manual insect counts and flower visits in raw / compressed videos.} ``Raw Video" presents observations from raw videos of \cite{Ratnayake2022_dataset}. ``Comp. video" presents observations from our compressed videos. ``Insects Missed" and ``Visits Missed" counts insect appearances and flower visits unobserved in compressed videos, but counted from raw video observations. ``New Insects Observed" and ``New Visits Observed" shows insect appearances and flower visits observed in compressed videos unrecorded in observations made on the raw video dataset. Counts related to ``New Insects Observed" and ``New Visits Observed" are included in ``Comp. video" recordings.}
\small
\label{tab:behaviour_analysis}
\begin{tblr}{
  width = \linewidth,
  colspec = {Q[134]Q[71]Q[83]Q[90]Q[146]Q[25]Q[71]Q[83]Q[88]Q[132]},
  column{odd} = {c},
  column{2} = {c},
  column{4} = {c},
  column{8} = {c},
  column{10} = {c},
  cell{1}{1} = {r=2}{},
  cell{1}{2} = {c=4}{0.39\linewidth},
  cell{1}{7} = {c=4}{0.374\linewidth},
  hline{1,7} = {-}{},
  hline{2} = {2-5,7-10}{},
  hline{3} = {1-5,7-10}{},
}
{\textbf{Insect}\\\textbf{Type}} & \textbf{Insect Counts} &  &  &  &  & \textbf{No. of flower visits} &  &  & \\
 & {\textbf{Raw }\\\textbf{video}} & {\textbf{Comp.}\\\textbf{video}} & {\textbf{Insects}\\\textbf{Missed}} & {\textbf{New Insects}\\\textbf{Observed}} &  & {\textbf{Raw}\\\textbf{video}} & {\textbf{Comp.}\\\textbf{video}} & {\textbf{Visits}\\\textbf{Missed}} & {\textbf{New Visits~}\\\textbf{Observed}}\\
Honeybee & 20 & 20 & 0 & 0 &  & 67 & 67 & 2 & 2\\
Syrphidae & 6 & 9 & 0 & 3 &  & 5 & 6 & 0 & 1\\
Lepidoptera & 4 & 5 & 0 & 1 &  & 6 & 6 & 0 & 0\\
Vespidae & 10 & 13 & 0 & 3 &  & 0 & 0 & 0 & 0
\end{tblr}
\end{table*}

The compressed video preserved all insect count information. These insects, primarily small-bodied members of the Syrphid, Lepidoptera, and Vespidae families, often visually blend into the environment, leading to them being missed by humans in unprocessed, raw video sequences where they are viewed against complex backgrounds (see Discussion). In comparison, the appearances of these insects were noticeable due to colour highlights against blacked-out surroundings in the compressed video. The compressed video captured all flower visits except for those by honeybees. This discrepancy occurred because the compression algorithm discarded frames with no motion resulting in missing frames where insects, particularly honeybees, decelerate rapidly and hover before landing on a flower, causing their motion to appear zero. %However, it's important to note that these counts were based on specific guidelines for identifying insect motion events constituting a flower visit.

\subsubsection{Automated video observations}

\begin{figure*}[h!]
    \centering
    \begin{subfigure}{0.925\textwidth}
    \centering
        \includegraphics[width=0.90\textwidth]{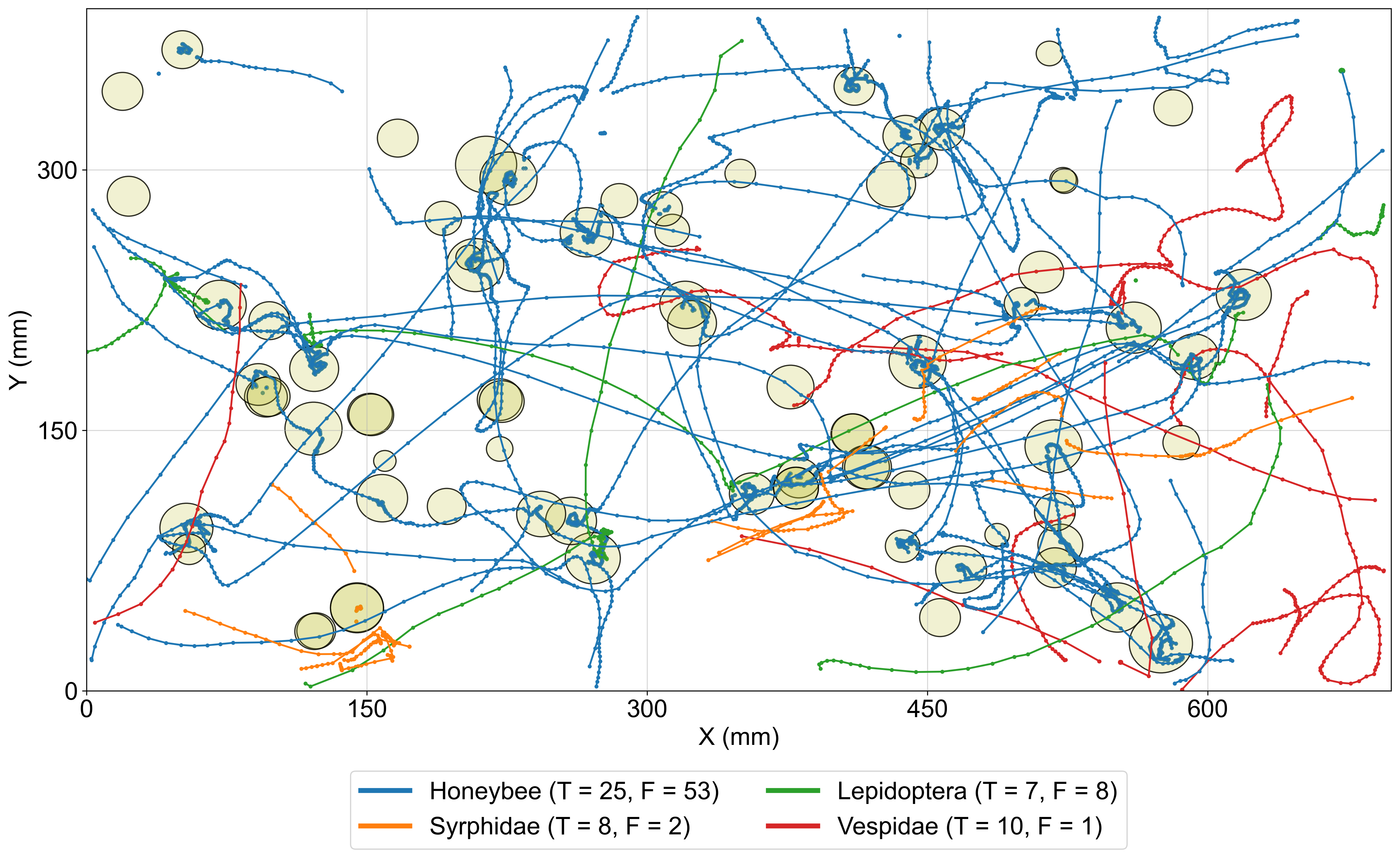}
        \label{fig:raw_tracks}
        \caption{}
    \end{subfigure}
    \hfill
    \begin{subfigure}{0.925\textwidth}
    \centering
        \includegraphics[width=0.90\textwidth]{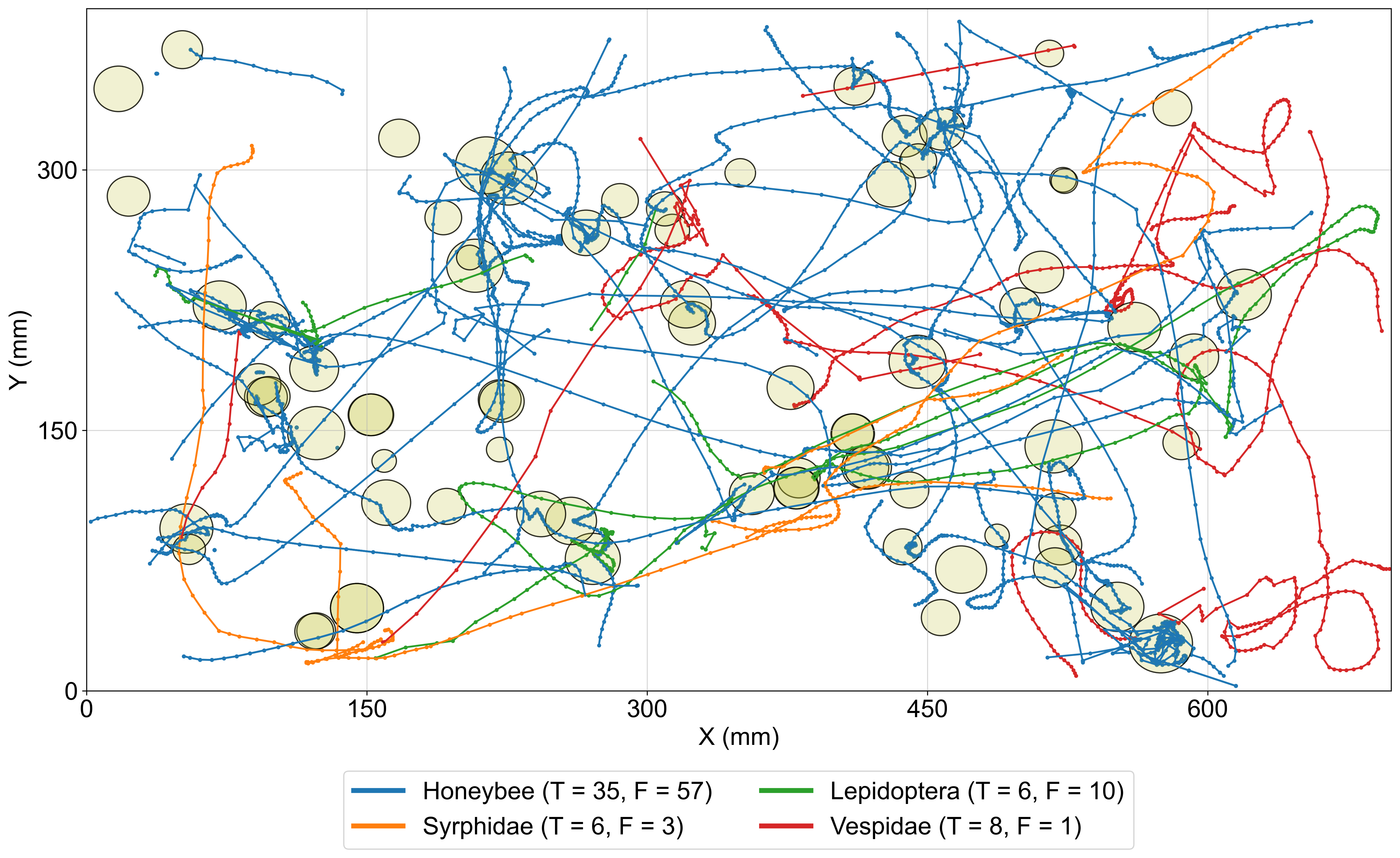}
        \label{fig:compressed_tracks}
        \caption{}
    \end{subfigure}
    \hfill
    \caption{\textbf{Insect trajectories and flower positions were extracted using automated video observations from (a) raw videos and (b) compressed videos.} Video processing was conducted using Polytrack \citep{ratnayake2023spatial}.  In the legend, for each type of insect, ``T" indicates the number of tracks recorded, ``F" denotes the number of flower visits made Flower locations are highlighted with yellow circles.}
    \label{fig:polytack_tracks}
\end{figure*}

To evaluate the suitability of the compressed videos for automated insect tracking and behavioural analysis, we used the modified Polytrack software (Section 2.2) to extract and compare insect trajectories and flower positions from raw and compressed videos. Extracted tracks were then post-processed manually by browsing through insect images associated with each saved track to correct insect type identifications and remove tracks originating from detections that do not correspond to insects. The results are plotted as insect trajectories in Fig.~\ref{fig:polytack_tracks}.

\clearpage

\subsection{Edge-device performance}

% \subsubsection{Video processing time}

We evaluated the performance of the proposed video compression algorithm on three edge computing platforms used for camera traps and insect pollinator monitoring: general purpose Raspberry Pi models 4 and 5 (8 GB), and a Nvidia Jetson Nano (2 GB). A detailed devices specification is provided in Supplementary information.

\begin{figure*}[h!]
    \centering
    \begin{subfigure}{0.49\textwidth}
        \includegraphics[width=\linewidth]{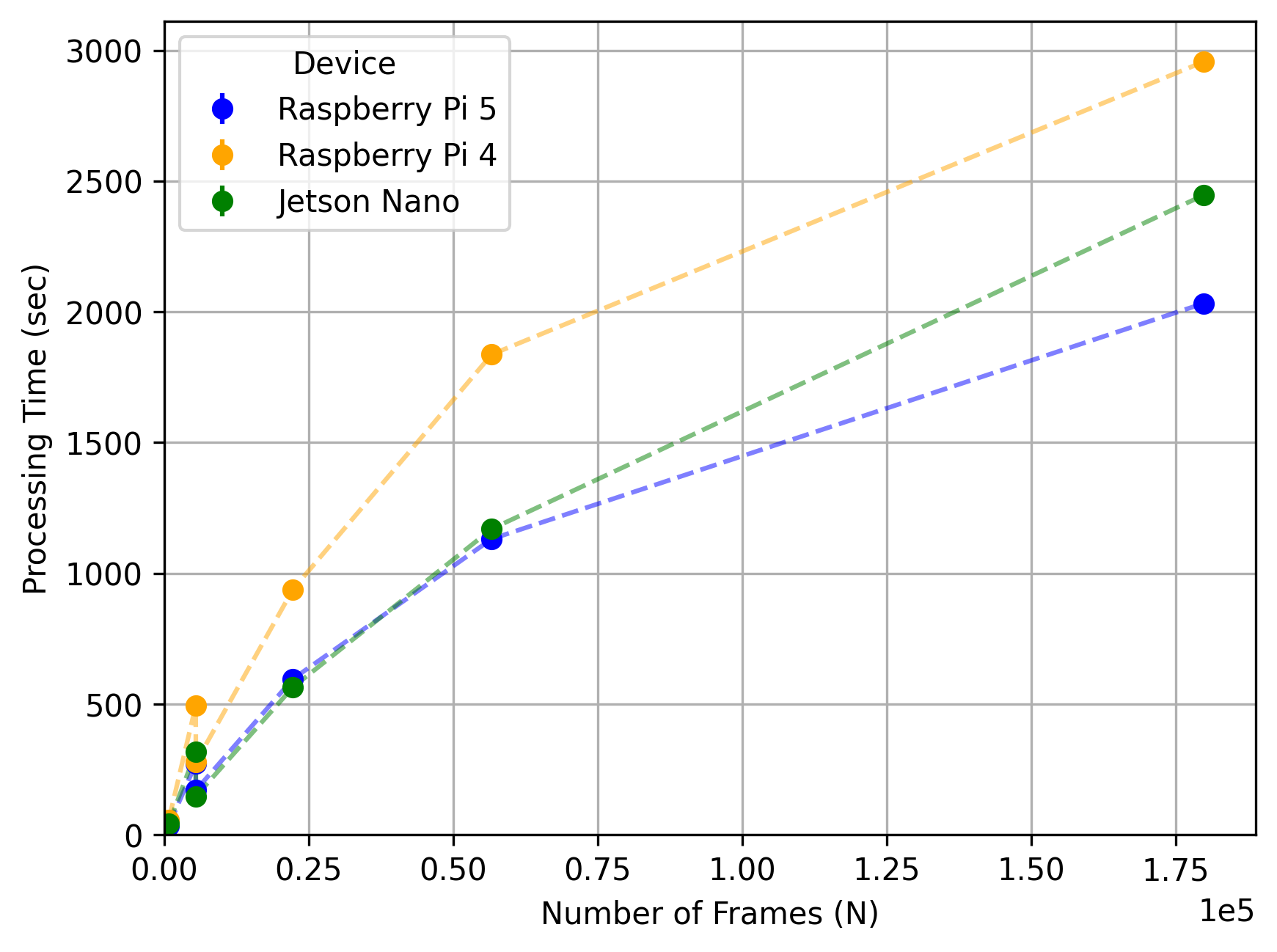}
        \label{fig:processing_time}
        \caption{}
    \end{subfigure}
    \hfill
    \begin{subfigure}{0.49\textwidth}
        \includegraphics[width=\linewidth]{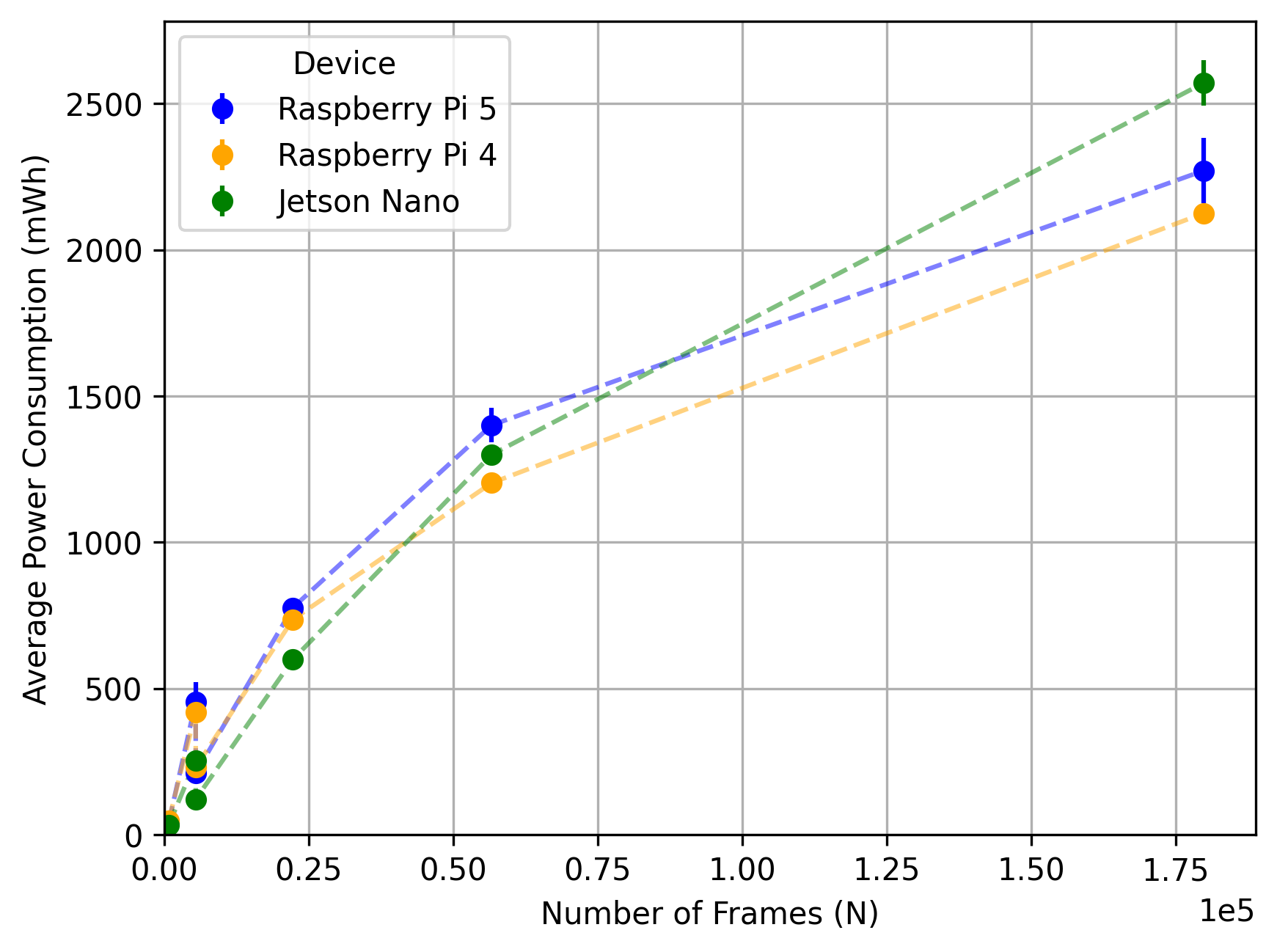}
        \label{fig:power_consumption}
        \caption{}
    \end{subfigure}
    \hfill
    \caption{\textbf{Performance of the proposed algorithm on edge processing devices.} (a) Average Processing Time and (b) Average Power Consumption of the proposed algorithm on various edge device platforms when processing the test datasets. Each dataset was processed in three replicates. x-axis shows the number of frames in the test dataset and error bars show the standard deviation of the mean.}
    \label{fig:edge_device_results}
\end{figure*}

Processing time was calculated by measuring the time taken by each platform to continuously process each video dataset. A USB power meter connected to the devices measured the energy consumed by each processing task. Experiments were conducted in a temperature-controlled room with the temperature varying between $20 \circ C - 23 \circ C$ Celsius. The power consumed by the video processing task was determined by removing the idle power consumption of the device from the power consumption measured by the USB power meter. Experiments were conducted for three replicates. Results are shown in Table \ref{tab:processing_time} and Fig.~\ref{fig:edge_device_results}.

\begin{table*}[h]
\centering
\caption{\textbf{Processing time for test video dataset on edge computing platforms.} ``Dur. (sec)" and ``FPS" show the total video duration and average video frame-rate for raw videos. ``Time (sec)" and ``Speed (fps)" present the processing time (Mean $\pm$ Standard deviation)  and average number of frames processed per second on each platform, based on three replicates.}
\label{tab:processing_time}
\small
\resizebox{\linewidth}{!}{%
\begin{tblr}{
  width = \linewidth,
  colspec = {Q[l,3cm]Q[c,1.5cm]Q[c,1cm]Q[c,2.5cm]Q[c,1.5cm]Q[c,2.5cm]Q[c,1.5cm]Q[c,2.5cm]Q[c,1.5cm]},
  cells = {c},
  cell{1}{1} = {r=2}{},
  cell{1}{2} = {c=2}{},
  cell{1}{4} = {c=2}{},
  cell{1}{6} = {c=2}{},
  cell{1}{8} = {c=2}{},
  hline{1,3,9} = {-}{},
  hline{2} = {2-9}{},
}
\textbf{Dataset} & \textbf{Raw Videos} &  & \textbf{Raspberry Pi 5} &  & \textbf{Raspberry Pi 4} &  & \textbf{Jetson Nano} & \\
 & {\textbf{Dur.}\\\textbf{(sec)}} & \textbf{FPS} & \textbf{Time (sec)} & {\textbf{Speed}\\\textbf{(fps)}} & \textbf{Time (sec)} & {\textbf{Speed}\\\textbf{(fps)}} & \textbf{Time (sec)} & {\textbf{Speed}\\\textbf{(fps)}}\\
\cite{naqvi2022camera} & 181 & 30 & $272.1 \pm 0.7$ & 20.1 & $493.7 \pm 2.1$ & 11.0 & $316.4 \pm 6.2$ & 17.2\\
\cite{Ratnayake2020_dataset} & 371 & 60 & $596.1 \pm 0.7$ & 37.4 & $938.4 \pm 1.2$ & 23.7 & $564.3 \pm 0.9$ & 39.5\\
\cite{Ratnayake2022_dataset} & 5597 & 30 & $2032.5 \pm 6.1$ & 88.5 & $2956.6 \pm 7.0$ & 60.9 & $2446.6 \pm 1.7$ & 73.5\\
\cite{van2022continuous} & 33 & 24 & $35.5 \pm 0.3$ & 22.3 & $57.7 \pm 0.0$ & 13.7 & $41.7 \pm 0.0$ & 19.0\\
Nest Monitoring & 2266 & 25 & $1131.2 \pm 6.9$ & 50.1 & $1837.0 \pm 5.1$ & 30.9 & $1169.6 \pm 3.0$ & 48.5\\
\cite{droissart2021pict} & 307 & 18 & $170.6 \pm 0.5$ & 32.1 & $277.0 \pm 0.7$ & 19.8 & $147.6 \pm 0.4$ & 37.0
\end{tblr}}
\end{table*}

The Raspberry Pi 5 outperformed the other two platforms on most test video datasets, achieving faster processing speeds and shorter processing times. Notably, the Jetson Nano was faster for the specific case of the \cite{droissart2021pict} dataset. Raspberry Pi 4 displayed slow processing speeds for all datasets. Jetson Nano was power efficient for processing smaller video datasets. However, when handling larger datasets like \cite{Ratnayake2022_dataset} the Raspberry Pi 4 was the most energy-efficient option.

\section{Discussion}\label{sec4}

% Even if manually observed, okay since only xx times to observe.
% Camera trap videos of animals captured in remote locations are invaluable for monitoring and understanding animal behaviour, including that of insects [REF again]. To address the resource constraints of camera traps, this study presented a novel video compression algorithm based on motion analysis. Our algorithm analyses pixel-wise motion within video frames using a multi-threaded approach to remove frames and individual pixels devoid of valuable information while retaining critical data on animal behaviour. In this section, we discuss the implications of our results.

\subsection{Video compression}

Our algorithm achieved over 87\% compression of insect monitoring videos across diverse monitoring environments (Table \ref{tab:video_compression}), while preserving key information (Table \ref{tab:behaviour_analysis}). This translates to reduced storage and bandwidth needs, particularly beneficial for resource-constrained edge-camera traps. The algorithm achieved file size reductions exceeding frame count reductions across all datasets. This was particularly pronounced in controlled environments with minimal background changes and for video files recorded with motion-based triggers or containing motion data in the majority of recorded video frames (e.g \cite{naqvi2022camera, van2022continuous, droissart2021pict}, Table \ref{tab:video_compression}). In these cases, our pixel-wise motion analysis selectively eliminated data from non-moving pixels while retaining essential information from pixels with motion for further analysis. This demonstrates the adaptability of our approach in compressing videos, regardless of the environmental conditions, the number of animals in the frame, or the type of trigger used for recording. 

Our approach utilises foreground changes as motion triggers without assessing cause of motion. While this approach can record animals with few false positives (Table \ref{tab:behaviour_analysis}), compressed videos did contain false (non-insect) detections caused by wind and illumination changes. These add to the compressed video file size, especially in dynamic monitoring environments. One potential strategy to minimise false detections involves assessing foreground change-based motion triggers using deep learning prior to storing the video frames \citep{zualkernan2022iot, tan2022animal, bjerge2021real, sittinger2023insect}. While this may improve compression, the effectiveness of the camera trap then hinges on the accuracy of the deep learning which, if inadequately trained, may miss events \citep{tan2022animal}. This is an issue especially for scarce or undocumented animal species that might be extremely valuable to record \citep{van2022emerging}. In addition, deep learning demands greater computational resources and more specialised, relatively expensive hardware than our approach, for it to process images on a device \citep{sittinger2023insect}. Deep learning may also increase frame processing latency, reducing capture rates and increasing the likelihood of missing fast animals. The efficacy of camera traps relies on their ability to capture a comprehensive representation of animals and their environment. Unforeseen animal interactions, new species, and previously undocumented behaviours may be overlooked if the camera is only programmed to capture data based on existing information. Therefore, it can be crucial to collect animal data unaligned with past (training) data. Furthermore, the complexities associated with animal behaviour may surpass the capabilities of current artificial intelligence models that can only be trained to identify a selected number of behaviours \citep{arablouei2023animal}. A complete raw record of animal appearances and behaviours though, allows experts to analyse videos beyond AI's current capability. The emergence of zero-shot object detection models shows promise in addressing the limitations of existing AI models \citep{bansal2018zero}. Therefore, future research on their application in camera trap-based animal studies would be valuable.

\subsection{Information retention}

In our manual comparisons of compressed and raw videos, videos processed by the algorithm successfully retained all information required for counting insects (Table \ref{tab:behaviour_analysis}). Additionally, our human observations of compressed videos revealed more fast-moving Vespids, relatively small Syrphids, and Lepidoptera than did the raw video observations. By discarding frames and pixels lacking motion activity and preserving colour information in pixels with motion, our algorithm removes the distraction of complex images where insects are only a tiny proportion of visible pixels, allowing humans to accurately identify insects. Conversely, video playback observations with raw videos require careful study of the entire complex frame for extended periods, often with little or no insect activity visible. Human attention limitations \citep{simons1999gorillas}, fatigue \citep{zett2022inter, faber2012mental}, and even low skill levels \citep{zett2022inter} then led to missed insect-related events. Discarding entire frames without motion activity clearly reduces the time, likelihood of errors, and ultimately the costs of ecological video observation \citep{breeze2021pollinator}.

Increasingly, automation of animal behaviour studies \citep{marks2022deep, schindler2021identification}, including outdoor insect tracking \citep{gebauer2024towards, ratnayake2021towards, ratnayake2023spatial, sittinger2024insect}, has enhanced our understanding of animal behaviour. Algorithms can independently process hours of video data, extracting animal movements and environmental interactions with little human intervention. %To enable the use of compressed videos for automated processing, we proposed a pre-processing step for the outdoor insect tracking Polytrack software \citep{ratnayake2023spatial} (Fig.\ref{fig:polytrack_overview}).

Polytrack, with its new pre-processing step, extracted most of the insect tracks and flower visits from the compressed videos that were extracted from the raw videos (Fig.\ref{fig:polytack_tracks}). This demonstrates the feasibility of automated tracking and behaviour analysis using compressed videos. However, some tracks were fragmented into segments during this process. Flower visit counts extracted from compressed videos were consequently higher than those extracted from raw videos due to track fragmentation. The Polytrack software used in this experiment was designed to process continuous videos without frame or video segment omissions. This limited its ability to handle omitted frames in compressed videos such as those caused by insects remaining stationary on flowers. Future research to design tracking methods for temporally inconsistent spatial data would be valuable. Advancements in the individual identification of animals will also enhance tracking performance by linking separate tracks either side of animal occlusions. Ultimately these improvements will lead to more robust and reliable animal tracking with compressed videos.

\subsection{Edge device performance}

The edge device performance evaluation (Table \ref{tab:processing_time}) shows processing speeds over 20 fps on all datasets on the Raspberry Pi 5. This was achieved by multi-threading processes for reading, processing, and writing video frames. Processing speed increased in semi-controlled environments with sparse insect appearances (\cite{Ratnayake2022_dataset} and Nest Monitoring datasets), regardless of the percentage of pixel changes per frame (Fig.\ref{fig:pixel_change}). This indicates that our algorithm has the potential for real-time video recording and compression. However, the algorithm's performance depends on the application environment, weather and device specifications. With increased insect activity, the camera-trap hardware might reach its computational limits, resulting in frame drops and missed fast-moving insects. Therefore, while real-time video recording is feasible in some scenarios, compressing videos through post-processing remains more reliable if it is possible. We anticipate that the development of more energy-efficient and powerful hardware will eventually enable the real-time implementation of our method across broader monitoring conditions, as evidenced by the increase in processing speeds from Raspberry Pi model 4 to 5 (Table \ref{tab:processing_time} and Fig. Fig.\ref{fig:edge_device_results}a).

As dataset size increased, the compression algorithm consumed more power across all edge device platforms (Fig.\ref{fig:edge_device_results}b). This adds to the overall monitor energy usage that includes video recording energy consumption. However, video compression can subsequently reduce the energy consumption and requirements associated with video transmission and storage by decreasing file sizes (Table \ref{tab:video_compression}). This is particularly advantageous for autonomous camera traps in remote, poorly connected locations. The compression algorithm's value will vary depending on the relative availability of energy, storage, and transmission resources in a scenario. Its additional power consumption might be met by solar power generation for instance in some circumstances. Advances in energy-efficient edge hardware may also lower power consumption, further expanding its valuable range.

. 

% As the size of the dataset increased, the compression algorithm consumed more power on all edge device platforms (Fig.\ref{fig:edge_device_results}b). This increase in power consumption contributes to the overall energy usage in addition to the energy consumed recording the video. However, it also leads to a significant reduction in file size of over 70\% for all test datasets (see Table \ref{tab:video_compression}). This reduction in size is advantageous for camera trap devices deployed in remote locations with limited storage and bandwidth. Despite the additional power consumption during processing, the size reduction provides a substantial benefit. The proposed methods can be further improved by advancements in energy-efficient edge devices, which can decrease power consumption and expand the range of applications.

\section{Conclusions}\label{sec5}

This paper presented a motion-analysis algorithm for compressing data captured by resource-constrained video camera traps. By analysing video frames pixel-by-pixel, our algorithm achieved an average compression of 87\% on diverse test datasets while preserving all information crucial for animal behaviour analysis. This substantial file-size reduction significantly enhances the monitoring capabilities of remote camera traps by maximising monitoring duration, minimising storage requirements and bandwidth demands, and facilitating efficient data transfer from resource-limited devices. In addition, our compression algorithm improved human observer effectiveness when extracting ecological data from videos. It directed observer attention to video sequences and frame regions with animal activity by removing sequences and pixels lacking relevant information. This alleviates human strain and fatigue associated with lengthy observation sessions. These reductions in camera trap resource requirements and observer time commitment have the potential to significantly reduce the cost of ecological monitoring. Hence, deploying this solution on camera traps can significantly advance ecological monitoring by improving the capabilities of existing systems.

\backmatter

\bmhead{Supplementary information}

Additional and detailed experimental results on presented methods are available in the Supplementary Information file ``Supplementary\_Information.pdf''.

% If your article has accompanying supplementary file/s please state so here. 

% Authors reporting data from electrophoretic gels and blots should supply the full unprocessed scans for key as part of their Supplementary information. This may be requested by the editorial team/s if it is missing.

% Please refer to Journal-level guidance for any specific requirements.

\bmhead{Acknowledgements}

Authors acknowledge Professor James Cook and his research team for sharing the videos of Nest Monitoring for experiments.

\section*{Declarations}

\begin{itemize}
\item Funding: This research was supported partially by the Australian Government through the ARC’s Linkage Projects funding scheme (project LP210200213). Lex Gallon was supported by a scholarship from the Dept. of Data Science and AI, Faculty of Information Technology, Monash University.
\item Conflict of interest: The authors have no competing interests to declare that are relevant to the content of this article.
\item Ethics approval and consent to participate: Not applicable.
\item Consent for publication: Not applicable.
\item Data availability: The datasets generated and/or analysed during this study are included in the Supplementary Information and are available on this  \href{https://drive.google.com/drive/folders/1UtuAxcJCP8hf6F3cFuzMcO9FeKCxv-kj?usp=share_link}{Google Drive}. The data stored in the Google Drive will be uploaded to a permanent repository, and a DOI will be provided upon acceptance of the manuscript.
\item Code availability: \begin{itemize}
    \item EcoMotionZip: \\ \href{https://github.com/malikaratnayake/EcoMotionZip}{github.com/malikaratnayake/EcoMotionZip}
    \item Polytrack: \\ \href{https://github.com/malikaratnayake/Polytrack}{github.com/malikaratnayake/Polytrack}
\end{itemize}
\item Authors' contributions: Conceptualization: Malika Nisal Ratnayake, Lex Gallon, Adel N Toosi, Alan Dorin; Data curation: Malika Nisal Ratnayake, Lex Gallon; Formal analysis: Malika Nisal Ratnayake, Lex Gallon; Funding acquisition: Alan Dorin; Investigation: Malika Nisal Ratnayake, Lex Gallon, Adel N Toosi Alan Dorin; Methodology: Malika Nisal Ratnayake, Lex Gallon, Adel N Toosi Alan Dorin; Project administration: Adel N Toosi, Alan Dorin; Resources: Adel N Toosi, Alan Dorin; Software: Malika Nisal Ratnayake, Lex Gallon; Supervision: Malika Nisal Ratnayake, Adel N Toosi, Alan Dorin; Validation: Malika Nisal Ratnayake, Lex Gallon; Writing – original draft: Malika Nisal Ratnayake; Writing – review \& editing: Malika Nisal Ratnayake, Lex Gallon, Adel N Toosi, Alan Dorin.
\end{itemize}

\noindent

\bibliography{sn-bibliography}% common bib file
% \input sn-article.bbl

\end{document}